\documentclass[scads,technicalreport,accept,pdftex,moreauthors]{Definitions/scads} 

\firstpage{1} 
\pubvolume{1}
\issuenum{1}
\articlenumber{0}
\pubyear{2023}
\copyrightyear{2023}
\datereceived{ } 
\daterevised{ } 
\dateaccepted{ } 
\datepublished{ } 
\hreflink{https://doi.org/} 

\usepackage{ulem}
\usepackage{nopageno}
\usepackage{datetime}
\mmddyydate

\usepackage[indentfirst=false]{quoting}
\newenvironment{myquotation}{
\begin{quoting}
}
{
\end{quoting}
}

\def\gray#1{\textcolor{gray}{\textit{#1}}}

\Title{Where did you get that? \\Towards Summarization Attribution for Analysts\footnote{An earlier version of this report appeared in the \textit{SCADS 2023 Proceedings}.}}

\TitleCitation{Title}

\Author{Violet B, John M. Conroy$^{1}$, Sean Lynch$^{2}$, Danielle M, Neil P. Molino$^{1}$, Aaron Wiechmann$^{2}$, and Julia S. Yang$^{1}$}

\AuthorNames{Violet B, John M. Conroy, Sean Lynch, Danielle M, Neil P. Molino,  Aaron Wiechmann, and Julia S. Yang}

\AuthorCitation{Lastname, F.; Lastname, F.; Lastname, F.}

\address{%
$^{1}$ \quad IDA Center for Computing Sciences, Bowie, MD, USA; \{conroy,npmolin,jsyang\}@super.org\\
$^{2}$ \quad Laboratory for Analytic Sciences, Raleigh, NC; \{sclynch,awiechm\}@ncsu.edu
}

\abstract{
Analysts require attribution, as nothing can be reported without knowing the source of the information. In this paper, we will focus on automatic methods for attribution, linking each sentence in the summary to a portion of the source text, which may be in one or more documents. We explore using a hybrid summarization, i.e., an automatic paraphrase of an extractive summary, to ease attribution. We also use a custom topology to identify the proportion of different categories of attribution-related errors.}

\begin{document}


\section{Introduction}

Automatic text summarization methods have been studied for over sixty-five years \cite{10.1147/rd.22.0159} and have shown promise as a content selection and filtering tool. Evaluations from the National Institute of Standards (NIST), such as the Document Understanding Conferences (DUC) and Text Analysis Conferences (TAC), help scope the problem and advance techniques in text summarization. The methods developed in the time frame of the conferences 2001-2011 were largely extractive-based methods, i.e., algorithms for finding critical portions of one or more documents, typically sentences, which best served the user's need for a summary.

In the last six years, significant progress in abstractive summarization has been made with the advent of recurrent neural networks for abstractive summarization \cite{journals/corr/SeeLM17}, and then transformer models. The transformer model can encode or decode text to generate summaries \cite{liu-lapata-2019-text}  \cite{lewis-etal-2020-bart}. The limits of transfer learning to perform multiple natural language processing (NLP) tasks were extended by \cite{T5:10.5555/3455716.3455856} to tackle dozens of tasks. OpenAI extended such approaches \cite{chatgpt}, and many research laboratories produce fluent text and multi-modal models (including images, speech, etc.)

We focus on automatic methods for attribution, i.e., linking each sentence in the summary to a portion of the source text, which may be in one or more documents.
Given that large language models too often \emph{hallucinate}, i.e., make statements that are either not supported by the text or even contradicted by the text, the problem of automatic attribution methods is particularly daunting. 

Key to our approach is a fast extractive summarization method, \texttt{occams}~\cite{analytics2030030}, which produces extracts with strong content coverage. The \texttt{occams} summary, typically less than 1K tokens, is then paraphrased with an abstractive summarizer. In contrast to this \emph{hybrid} approach, the abstractive-only method requires applying the neural abstractive model to the entire set of input documents, increasing the computational cost and complicating the attribution problem.

Automatic methods can link summary sentences with likely attributing sentences in the source documents. To test the accuracy of these methods, we employ human evaluation. For each sentence in the abstract, one or more putative supporting sentences are presented to a human to judge if these sentences support, partially support, contradict, partially contradict, or are neutral. 

To test if the hybrid approach to summarization makes the problem of attribution easier, we compare it to an abstractive method on three datasets representing a range of information tasks for an analyst. We describe these in section \ref{sec:data}.

We note that the recent paper ``Attribute First, then Generate'' by \citet{slobodkin-etal-2024-attribute} takes an alternate approach of identifying relevant source segments before conditioning the generation process. This approach like our hybrid improves the citation accuracy. Also, \citet{worledge2024extractive} introduce the extractive-abstractive spectrum, with search engines and large language models as extremes with multiple intermediate approaches. The authors found ``...we find that perceived utility improves by as much as 200\%, while the proportion of properly cited sentences decreases by as much as 50\% and users take up to 3 times as long to verify cited information.''

\section{Summarization Algorithms}

In this experiment, we evaluate two automatic summary methods: abstractive and hybrid summarization. While both ways utilize GPT-3.5 Turbo (GPT in the rest of this paper), we use different inputs and prompts to change how the summaries are generated. The abstractive method will use GPT on the entire document to generate the final summary, relying on how the model itself was trained to handle summarizing documents. The hybrid approach will use \texttt{occams} to extract the most important text from the document. Then GPT will be used to rewrite the resulting paragraph into a more human-readable final summary. This section will outline how we execute each of these methods.

\subsection{Abstractive Summarization}

To generate our abstractive summaries, we will provide basic instructions for GPT. This experiment compares a hybrid summary with a purely abstractive summary, so we want to allow GPT to behave according to its training. While some recent research has investigated the impact of different prompting styles on the quality and factuality of summaries generated by large language models, we are seeking to evaluate the summaries that the model generates without refinement.

With these goals in mind, we used the following prompt to generate the abstractive summaries for all three datasets:

{\fontfamily{qcr}\selectfont
\noindent``````

\noindent You are an abstractive summarizer that follows the output pattern:

\noindent Text:
\noindent \{text\}

\noindent Summary:

\noindent ''''''
}

We specify an output pattern to ensure no extraneous references to the source text or other words beyond the expected summary. Also, it is essential to note that we do not constrict the length of the abstractive summary, allowing GPT to make it as long as it would by default. With that, we do generate these summaries early in the pipeline because the lengths of these summaries will influence the length we specify to \texttt{occams}.

One consideration we made when generating these summaries is the token length of our input. To prevent over-complicating our pipeline, we sought to abstractively summarize each document all at once in a single API call to GPT. While this was possible for the DUC/TAC and Cyber Threat Intelligence datasets, given the length of their documents, CrisisFACTS had some events with too many documents to process all at once. To reduce the number of documents for each event to a number that can be passed to GPT, we filtered the documents by their provided importance score and removed duplicates. This way, we could ensure that we had enough tokens to process the entire summary without sacrificing the most essential information.

\subsection{Hybrid Summarization}
We generate the hybrid summaries for each document via a two-step process, with the first step being carried out using the \texttt{occams} package while the second is a rewrite using GPT.

\subsubsection{OCCAMS Extractive Summarization}

Unlike GPT, \texttt{occams} requires a word or character budget to generate summaries. We need to use a different character budget for our experiment depending on the dataset we are working with. This is because we want to be able to compare the hybrid and abstractive summaries, and length is a major contributing factor when determining how much of the information from the source text can be represented within the summary.

We will determine this character budget for the DUC/TAC and Cyber Threat Intelligence datasets based on the abstractive summaries from GPT. Because these summaries are of varying lengths, we decided to set the budget at the 75th percentile of the length of all abstractive summaries for each dataset. We decided to go for this instead of the median length because \texttt{occams} is very rigid with its budget and will not exceed it. It will get very close to this target length for the summary, with minimal variation. There is a lot of variation within the abstractive summary lengths, and we must ensure that outliers will not impact the length we choose for our extractive summaries. 

For CrisisFACTS, the nature of the dataset does not require as much consideration for the length of the summary. Given that the discrete documents in the dataset tend to be much shorter, we expect that our extractive summary will pull individual documents to construct our fact summary. Additionally, our gold-standard summaries are already extractive and represent an ideal extractive summary. Consequently, we use the median summary from the facts to generate our extractive summaries.

Upon determining the character budget, we can generate the summaries for each dataset using \texttt{occams}. This will be directly used in the next step of our pipeline to determine our final hybrid summary.

\subsubsection{GPT Summarization}

We will use GPT to rewrite our abstractive summaries for the second step in our hybrid summarizer. Given that the advantages of abstractive summarization over extractive summarization are mostly related to readability and human preference, we decided to use a different prompt than that used for abstractive summaries. This is because we seek to rewrite our extractive summaries, generating a more cohesive paragraph rather than a collection of sentences. With that, we want to make sure that no additional information is being added or hallucinated by GPT. With this goal, we used the following prompt:

{\fontfamily{qcr}\selectfont
\noindent``````

\noindent Please rewrite the following into a coherent and readable paragraph. Do not deviate from the facts of these sentences or add any new information. Follow the output pattern:

\noindent Text:
\noindent \{text\}

\noindent Summary:

\noindent ''''''
}

Like the abstractive prompt, we provide an output format to ensure the generated summary has no extraneous words or sentences. We specify that no information should be added. It is also important to note that we are still not concerned with length and do not expect the final summary to be shorter than the extractive summary. The output returned by this prompt will be our final hybrid summary used for our evaluation.

\section{Three Exemplar Datasets}
\label{sec:data}
To sample a range of content, we performed our attribution study on three open datasets, which mimic specific analysts' needs. First, the CrisisFACTS data set\footnote{CrisisFACTS (https://crisisfacts.github.io) is a track at the NIST Text Retrieval Conference (TREC) meant to mimic the needs of crisis managers from the Federal Emergency Management Agency as they monitor an ongoing disaster, such as a fire or flood.} contains extracted content snippets (from Twitter, Reddit, and news articles) that are relevant to the current query. Second, a Cyber Threat Intelligence data set prepared by Elemendar for the Laboratory for Analytic Sciences (LAS) represents a specialist requirement, in this case, analysts who keep up with cyber threat activity. Third, the Text Analysis Conference (TAC) 2011 data set is a multi-document summarization data set of newswire documents where the documents are grouped by topic.

The CrisisFACTS dataset from the Text Retrieval Evaluation Conference (TREC) 2022 and 2023 is a good model of a task-based summary where an analyst, in this case, one working for FEMA, has a requirement to write a report on an ongoing disaster such as a fire or flood. For the 2022 dataset, there are eight events. Multiple report requests were made for these events, with the total number of reports (summaries) being 55. As some of the summaries and retrieved relevant documents were small, we filtered the EventID to remove those with too few facts or too many.\footnote{In particular, we filtered the data so that min\_sum\_len=200, max\_sum\_len=5000, min\_doc\_len=100, where length is measured in characters.} An EventID labels each report request. For each EventID, there is a list of facts, loosely an extracted summary for the period of the given EventID. Note there is an abstractive summary for each of the eight events from Wikipedia. We opted not to use these in our study.

The Cyber Threat Intelligence dataset was explicitly created for SCADS by Elemendar. These data contain reports on cyber events. Each document includes a summary. The document set consists of 59 training documents and 15 tests. We will sample the training data for our initial study. In addition to the documents, there are knowledge graphs and some sample questions and answers, which could be used for future content evaluation and fact-checking.

The Text Analysis Conference (TAC) 2011 dataset is a classic multi-document summarization task by the National Institute of Technology and Standards (NIST). The task of interest from 2011 is the guided summary task, which was to perform a multi-document summary of 10 documents on a given topic. The dataset has 45 such topics and an updated summary of each topic. For each topic, there are four human-written summaries.

\section{Automatic Attribution}

SummaC is by no means perfect for attribution, as experimentation indicates the metric has an underlying bias toward extractive summaries. During most of our experiments where we computed scores for abstractive, hybrid, and extractive summaries, SummaC consistently scored extractive summaries higher than any other type. We believe this is due to the Natural Language Inference (NLI) used initially in the metric pipeline to score the entailments of document-summary sentence pairs, as these probabilities are run through the model to output the score. 
After performing a document reduction experiment on every summary type where we ordered the document sentences by ascending neutrality, reduced the original document by removing the most neutral sentence, and recomputing the SummaC score, we found that only a few document sentences influenced the SummaC score for extractive summaries. Notably, the exact number of document sentences matching the total summary sentences contributed most heavily to the score, implying that source text facts are not evenly distributed among extractive summary sentences. On the other hand, for abstractive and hybrid summaries, more document sentences were relevant to the score. This indicated that source text facts are more evenly distributed for hybrid and abstractive. Because of this even distribution, no single summary sentence is highly entailed by a document sentence, so the entailment probabilities are typically low. By extension, the SummaC score is very low, but this output doesn’t necessarily mean the abstractive or hybrid summary is bad. Instead, SummaC is bad at capturing summary text's abstract nature. As a result, using solely SummaC as the attribution solution for an entire summary would cause potential unwanted disqualification of summaries from being marked as consistent with the original document. 

For this human evaluation experiment, we propose using NLI for sentence-level attribution as opposed to the SummaC metric for the entire summary. Because NLI produces low entailment probabilities for abstractive and hybrid summary sentences, we need to account for any missing context for a single abstracted summary sentence. To do this, we combine a targeted document sentence with a previous and following sentence to provide a range of context. Then we present this three-sentence text to our human evaluators for attribution labeling.

To test the ability of automatic methods for attribution, we use a natural language inference (NLI) model to find good matching sentences for each sentence in the GPT and Hybrid summaries. Following \cite{laban-etal-2022-summac}, we use the NLI \texttt{roberta-large-mnli} model. We also use a T5 sentence embedding cosine similarity for a baseline, with the \texttt{sentence-t5-xxl} model. 

\section{Human Labeling Experimental Design}
\label{sec:experiment}

Our evaluation methods are motivated by a recent paper \cite{liu2023evaluating}.
\section{Results}

The three summary types are human-written and two machine-generated summaries.
The machine-generated methods are purely abstractive using OpenAI GPT-3.5 Turbo model 
\texttt{gpt-3.5-turbo-16k-0613} (GPT) and hybrid, which is a paraphrase of an extractive summary generated by \texttt{occams}~\cite{analytics2030030} (hybrid). We note that the hybrid method is a paraphrased extract, and the extract's length was chosen to be comparable with the GPT summary to make the comparison easier. The input limit of the GPT model is 16K tokens, which generally allows GPT to summarize the entire document without repeatedly applying it. The only exception is the CrisisFACTS data.

\subsection{Content-Based}
We compare the two summarization approaches, GPT and hybrid, and benchmark their content coverage on the three datasets. Here, we measure the agreement of human-written summaries with the two automatically generated methods. To this end, we use both ROUGE \cite{lin-2004-rouge} and a recently proposed measure SMART \cite{amplayo2023smart} as a check to indicate how closely the automatic summaries match the human-written ones. Both these metrics correlate well with human assessment of content coverage. Table~\ref{tab:RS} gives the mean scores of these automatic evaluations. Broadly, the two summarization methods perform comparably regarding their content similarity with the human summaries for each dataset. Each dataset has about 50 documents (or document sets in the case of TAC 2011). In most instances, the differences are insignificant, meaning that both methods likely have comparable quality content.

\begin{table}
\begin{center}
\begin{tabular}{l|c|c|c|c}
Dataset &
\multicolumn{2}{c|}{GPT} &
\multicolumn{2}{c}{Hybrid} \\
& {ROUGE-2} & {SMART-2} & {ROUGE-2} & {SMART-2}\\
\hline
CrisisFACTS & 3.37 & 15.3 & 3.90 & 15.3  \\
Cyber Data & 10.2 & 24.9 & 9.3 & 22.9  \\
TAC 2011 & 11.5 & 24.4 & 10.4 & 24.5 \\
\end{tabular}
\end{center}
\caption{Automatic Evaluation Scores.} \label{tab:RS}
\end{table}

\subsection{Attribution}
We now turn to the question of automatic attribution. We evaluated two approaches for attribution, one based on an NLI model and the second using a T5 sentence embedding. We were uncertain how important the context would be for allowing the analysts to judge if the automatic attribution gave any support or contradicted the summary sentence. To this end, we considered two approaches when presenting the sentences in the document that the NLI or embedding model chooses for attribution. We called these two approaches Task 1 and 2 in the study.

\begin{figure}[H]
\centering
\includegraphics[height=5cm]{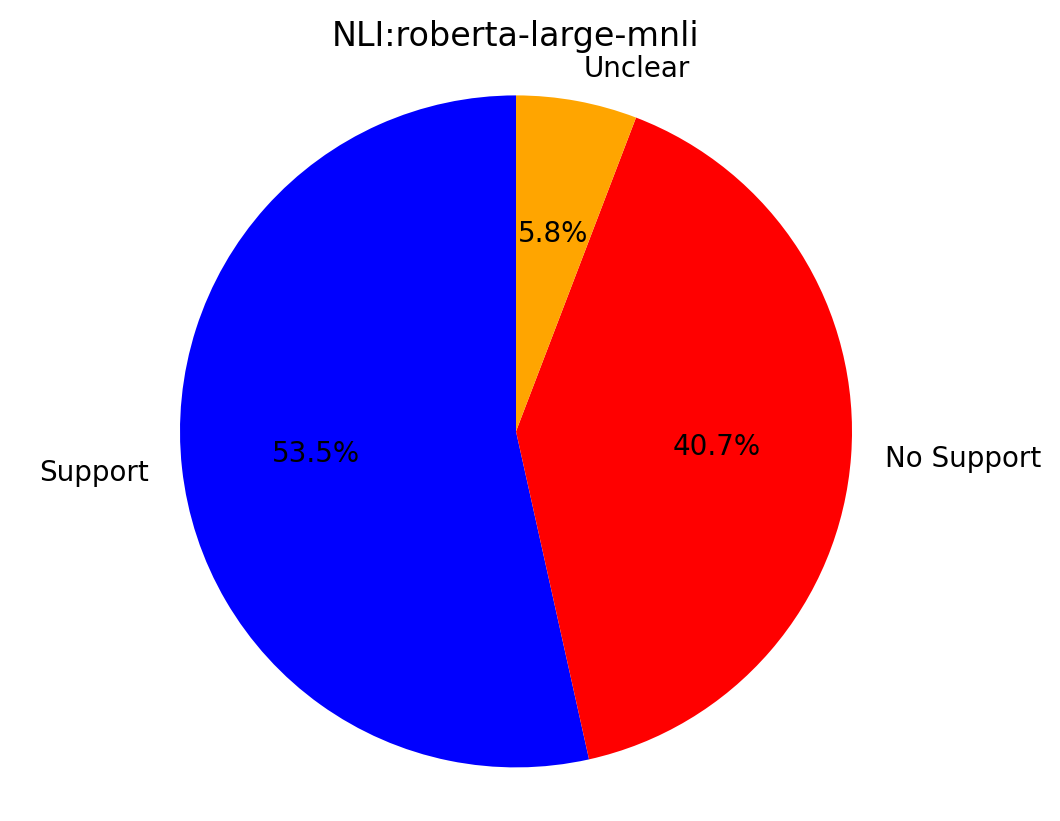}
\includegraphics[height=5cm]{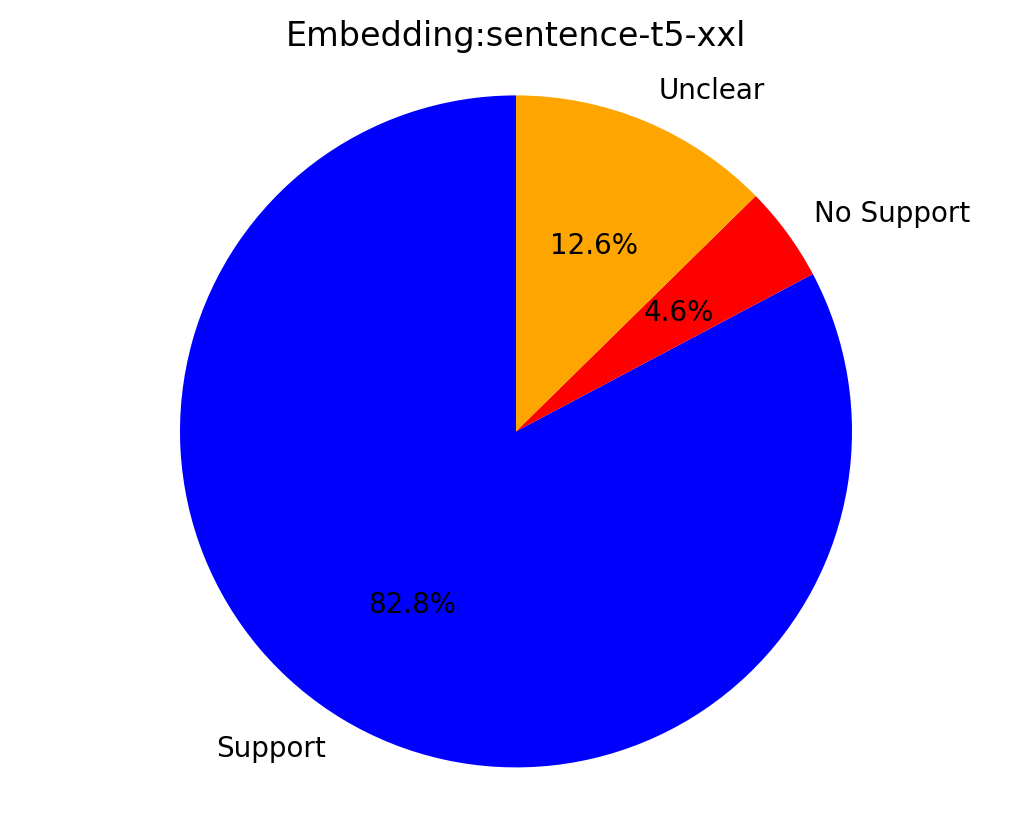}

\caption{Using embeddings in Task~1 gives stronger automatic attribution as judged by analysts compared to the NLI model.\label{Task1:human}}
\end{figure}

\begin{figure}[H]
\centering
\includegraphics[height=5cm]{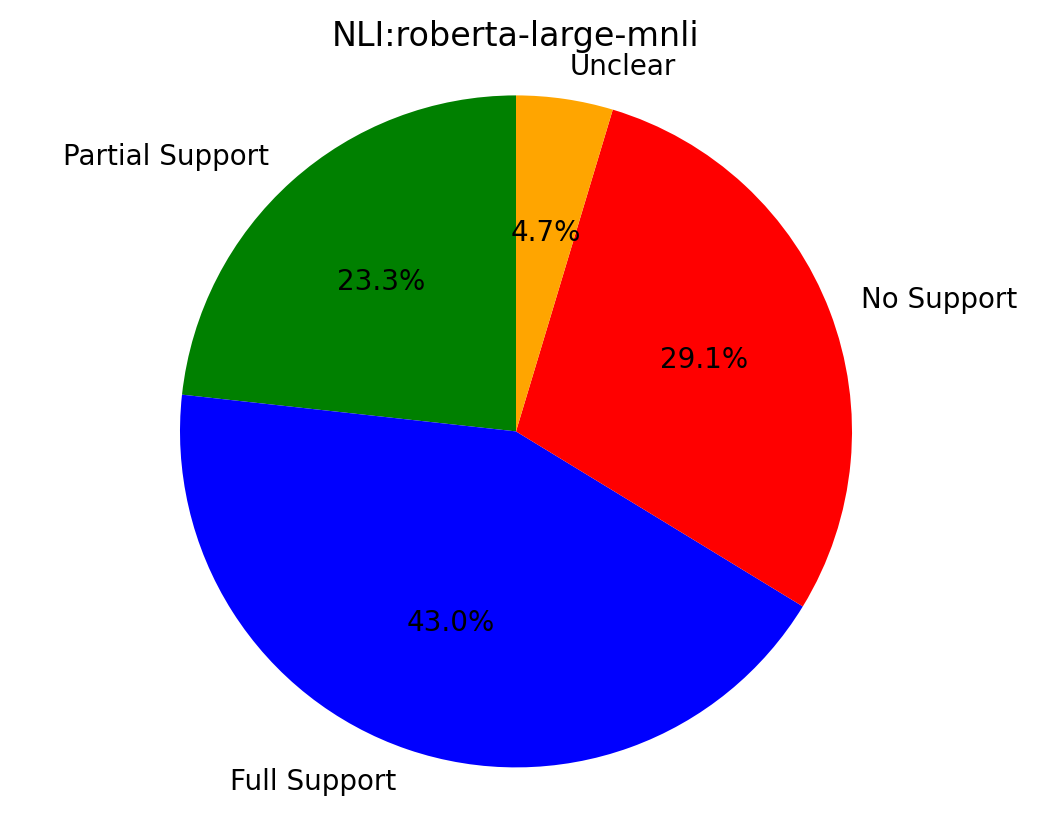}
\includegraphics[height=5cm]{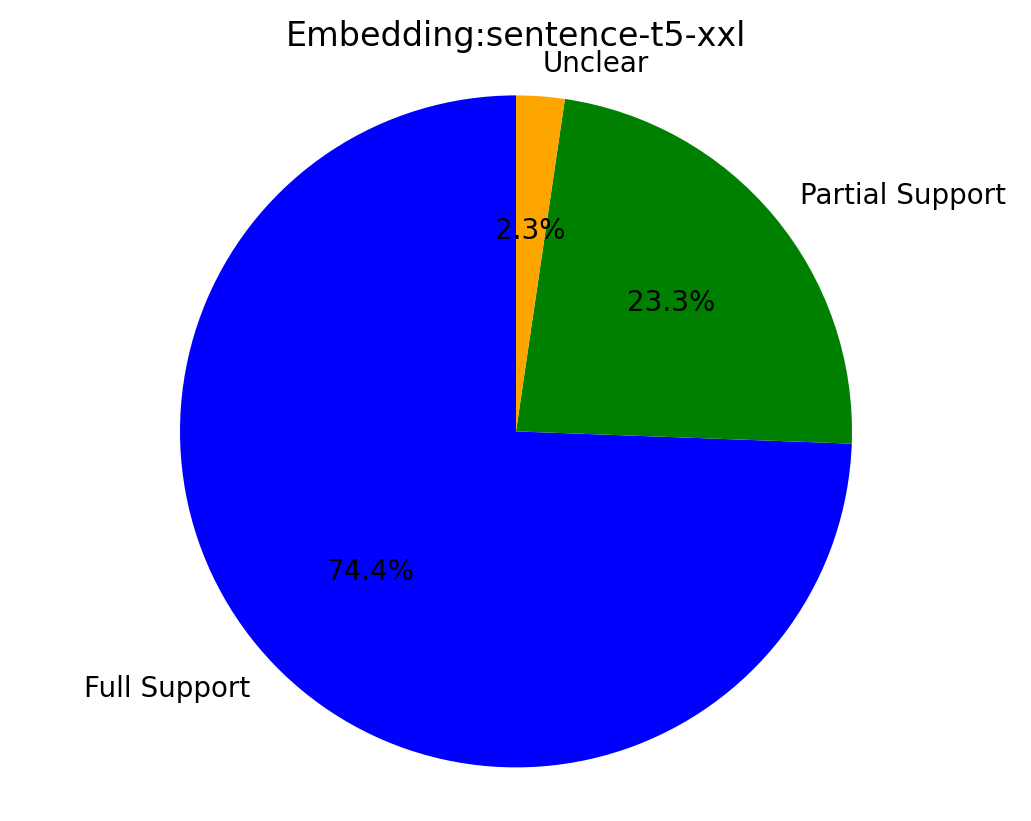}

\caption{Using embeddings in Task~2 gives stronger automatic attribution as judged by analysts compared to the NLI model.\label{Task2:human}}
\end{figure}

In Task 1, the attributing document sentence is presented in context with the sentence preceding and following it. In Task 2, we opted to allow the automatic attribution methods to give three supporting sentences, as some summary sentences may be a fusion of two or more sentences. Figures \ref{Task1:human} and \ref{Task2:human} give pie charts summarizing the results of the three analysts' annotations of 30 summary sentences to attribution text pairings for the NLI and embedding approaches. Here, the summaries are the human-generated summaries from TAC 2011. We note that in both tasks, the sentence embedding method gave attributions that the analysts judged better than the NLI model. Also, it appears that giving three possible supporting sentences gave more accurate attribution.

\begin{figure}[H]
\centering
\includegraphics[height=5cm]{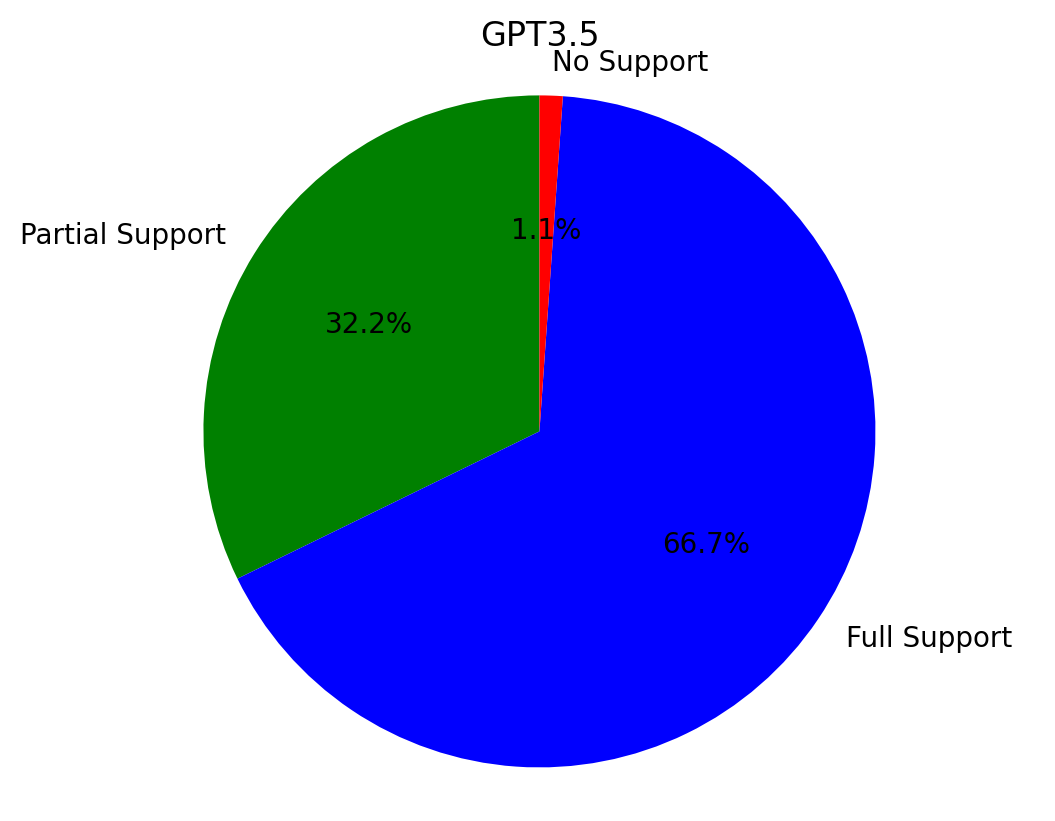}
\includegraphics[height=5cm]{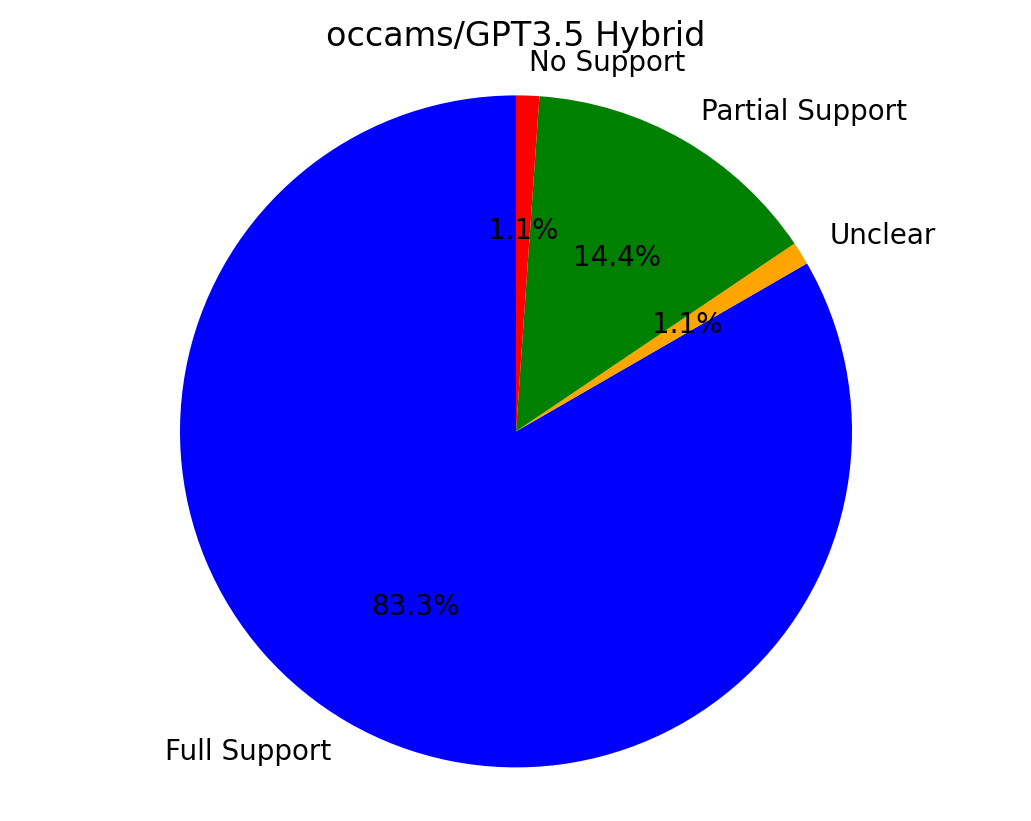}

\caption{Using Hybrid (\texttt{occams}/GPT) summaries on the TAC 2011 dataset gives rise to stronger automatic attribution as judged by analysts compared to using GPT alone.\label{Task2:machine}}
\end{figure}

We next compared summaries generated by GPT on the TAC 2011 data with a hybrid approach where we asked GPT to paraphrase \texttt{occams} extracts. To limit the amount of annotation, we decided only to use the embedding model and the Task 2 approach of providing the top three sentences for the automatic attribution evaluation. In Figure~\ref{Task2:machine}, the hybrid summaries were easier to attribute automatically. Note that the attribution sentences for the hybrid were selected from the extracted summary. As the summaries are approximately 100 words long, the analysts were given a significant portion of the original extract, typically about 5 sentences.

\begin{figure}[H]
\centering
\includegraphics[height=5cm]{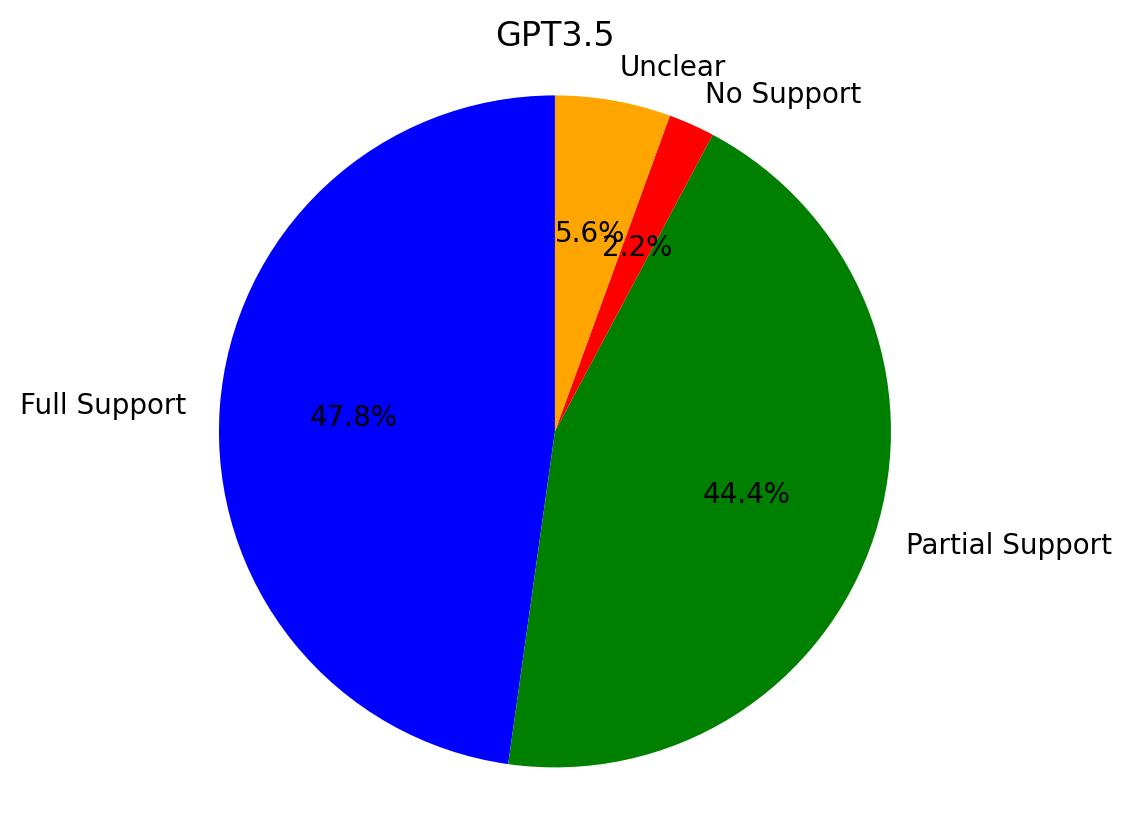}
\includegraphics[height=5cm]{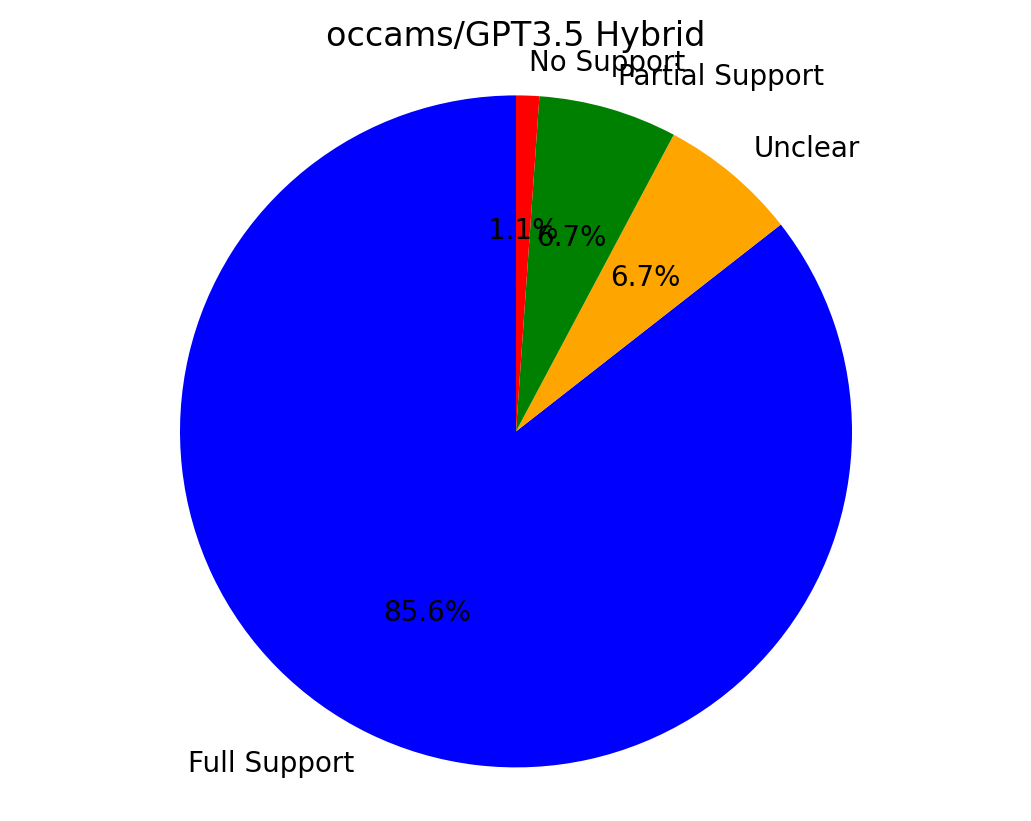}

\caption{Using Hybrid (\texttt{occams}/GPT) summaries on the Cyber dataset gives rise to stronger automatic attribution as judged by analysts compared to using GPT alone.\label{Cyber:analysts}}
\end{figure}

\begin{figure}[H]
\centering
\includegraphics[height=5cm]{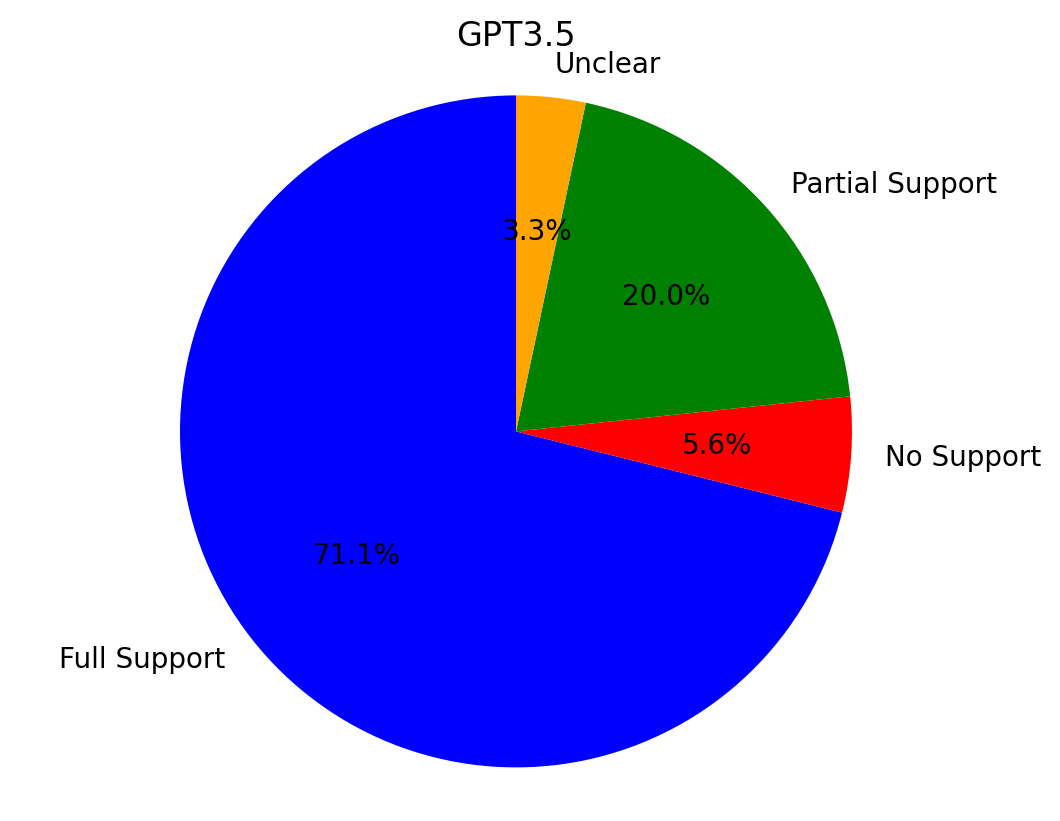}
\includegraphics[height=5cm]{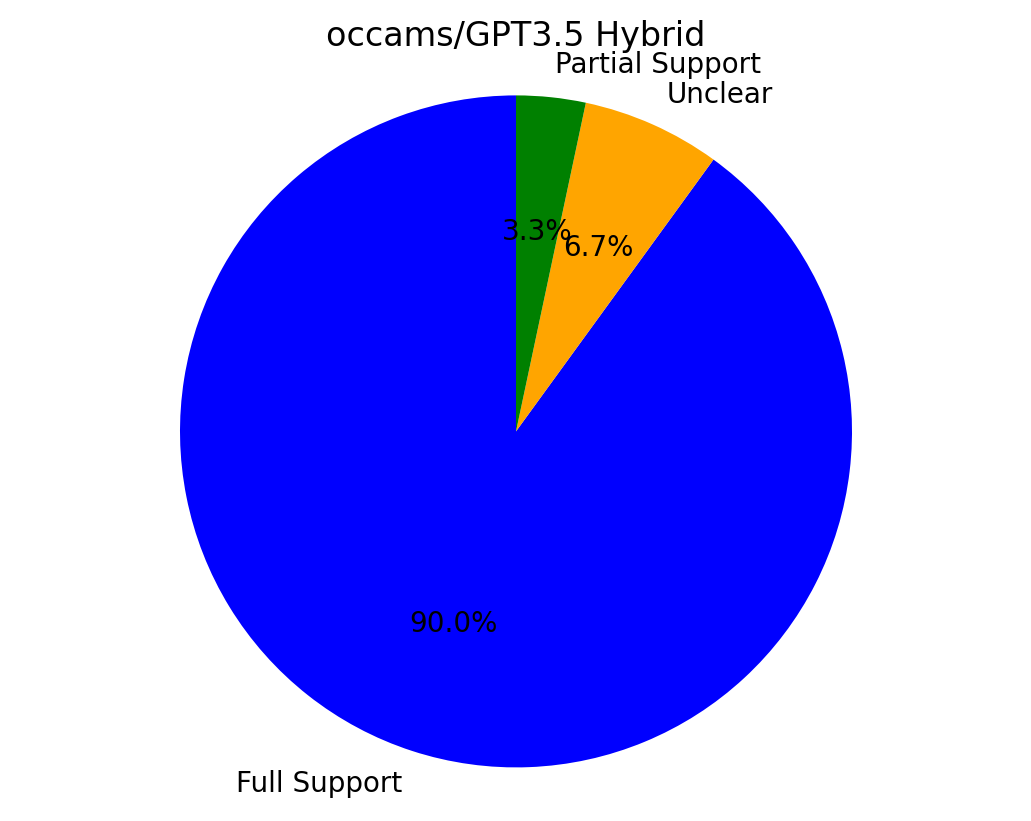}

\caption{Using Hybrid (\texttt{occams}/GPT) summaries on the Cyber dataset gives rise to stronger automatic attribution as judged by cyber experts compared to using GPT alone.\label{Cyber:experts}}
\end{figure}

For the Cyber Threat Intelligence data, we conducted a similar experiment. We generated GPT summaries and hybrid summaries (GPT paraphrases of \texttt{occams} extracts). As the Cyber data might be consumed by both a general audience and experts in cyber security, we separated the annotation into two groups. The first group with general knowledge is labeled as ``analysts,'' and the second group with expertise in cyber security is labeled as ``experts.'' Figures \ref{Cyber:analysts} and \ref{Cyber:experts} give the results. In both cases, the hybrid approach was more likely to provide attribution with full support. Of note is that the cyber experts judged the attributions to be of higher quality. Specifically, the cyber experts labeled the attributions as giving ``full support'' more often than analysts without a background in cyber security.

\begin{figure}[H]
\centering
\includegraphics[height=5cm]{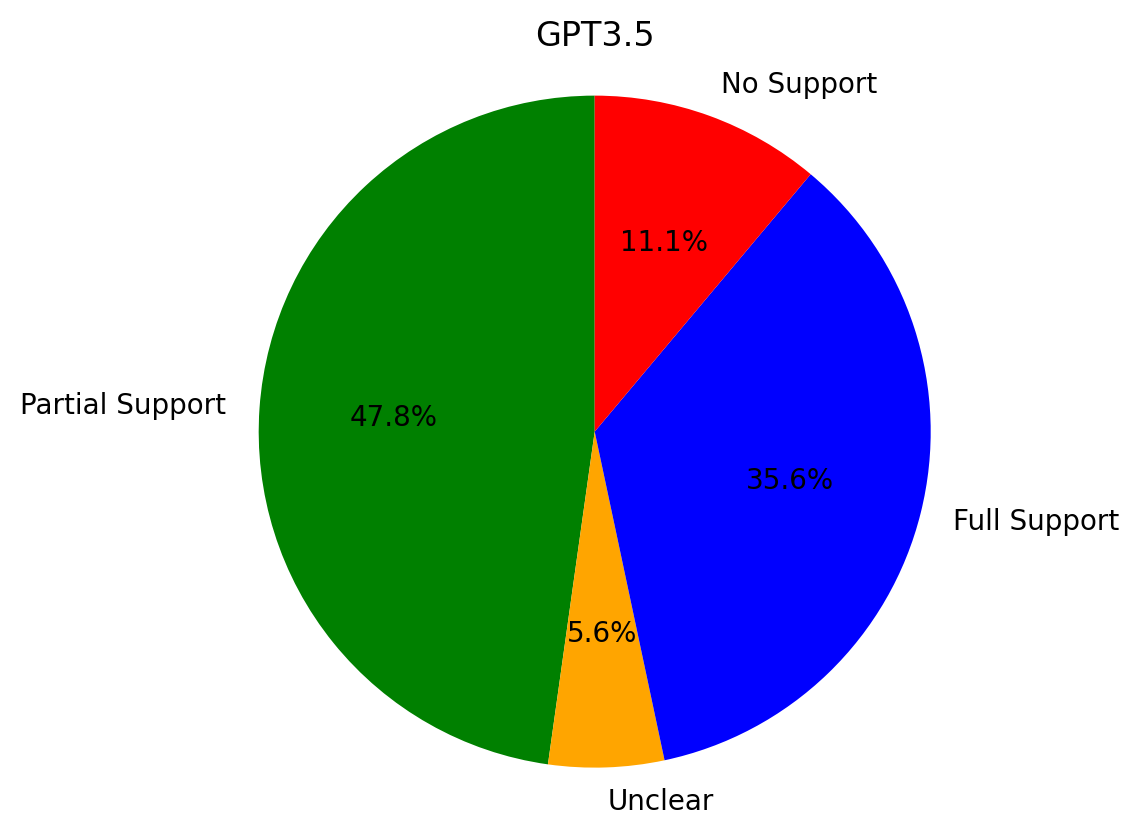}
\includegraphics[height=5cm]{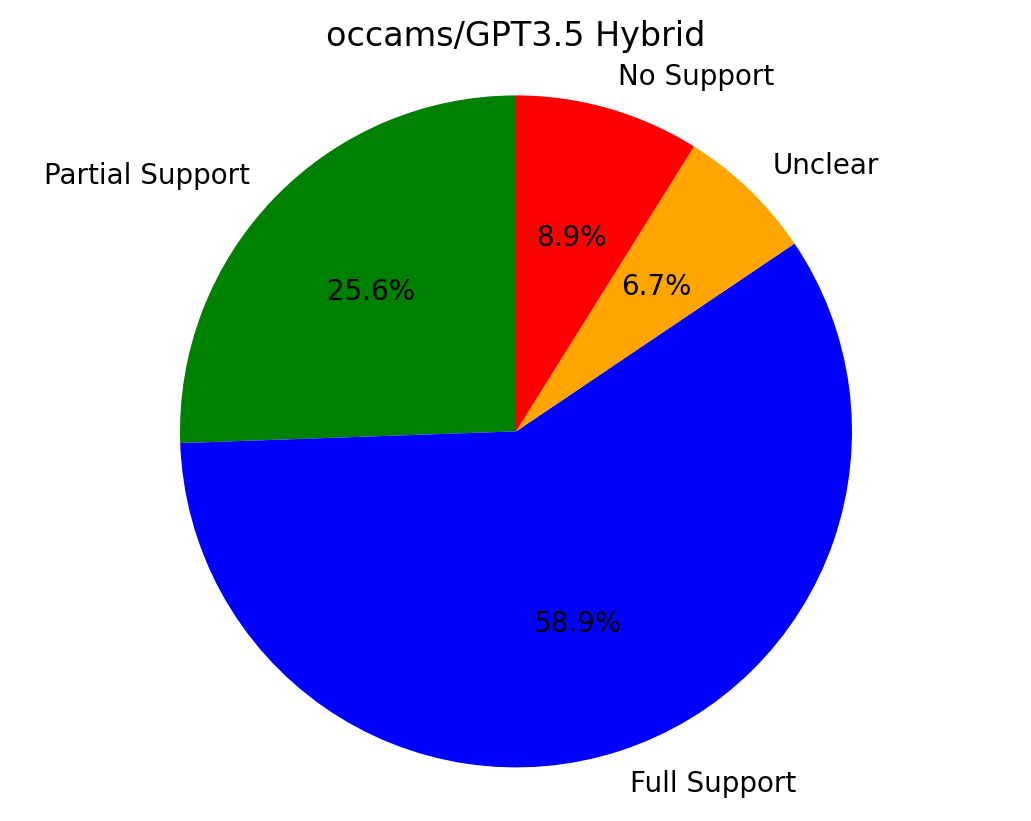}
\caption{Using Hybrid (\texttt{occams}/GPT) summaries on the CrisisFACTS dataset gives rise to stronger automatic attribution as judged by analysts compared to using GPT alone.\label{CrisisFACTS}}
\end{figure}

For the CrisisFACTS automatic attribution results (Figure~\ref{CrisisFACTS}), we see again the hybrid summary sentences had better automatic attribution than those using GPT without the \texttt{occams} extracts. As this data set was the least fluent of the three, it is not surprising to see that there was a drop in the fraction of attributions that received \emph{full support} relative to the results in Cyber and TAC datasets. For example, with the CrisisFACTS hybrid summaries, 58.9\% of the sentences had full support attributions. In contrast, with the TAC and Cyber datasets, the corresponding summaries had 74\% full support or more.

\subsection{Refutation}

Finally, we tally the percentage of refutations found in each experiment. Table~\ref{tab:refT12} gives the percentage of refutations for Task~1 and 2. Here, the embedding attribution method is more likely to link to a refuting sentence than the NLI approach. 

\begin{table}[H]
\centering
\begin{tabular}{c|c|c}
 &NLI  & Embedding \\
\hline
Task 1 & 0.0 & 10\\
Task 2  & 7.0  & 10\\
\end{tabular}
\caption{Percentage refutations for Task 1 and 2.}
\label{tab:refT12}
\end{table}

Table~\ref{tab:hybridVSgpt} compiles the remaining refutation results for the TAC and Cyber datasets. Here, the hybrid approach had fewer refutations than the GPT purely abstractive method except for the CrisisFACTS data where they both had about 8.9\% refutations.

\begin{table}[H]
\centering
\begin{tabular}{c|c|c}
 & Hybrid & GPT\\
\hline
TAC 2011   & 5.6 & 13.3\\
Cyber Analysts & 7.8 & 14.4 \\
Cyber Experts & 8.9 & 10.0 \\
CrisisFACTS & 8.9 & 8.9 \\
\end{tabular}
\caption{Percentages of refutations in all annotations comparing GPT and hybrid approaches.}
\label{tab:hybridVSgpt}
\end{table}

An in-depth analysis of the refutations follows.

\subsection{Refutation: In-Depth}\label{sec:refutation-analysis}

Three data sets -- TAC 2011 (multi-document over time), Cyber Threat Intelligence (single document, specialist) and CrisisFACTS (multi-stream, closest to raw data) -- were analyzed for different types of refutations. The findings were as follows:
\begin{itemize}
\item Allowing 3 attributing sentences instead of 1 improved support significantly, but has the potential to add sentences of lower quality.
\item Refutations often differ from LLM hallucinations in a machine summary.
\item The types of refutations reflect the underlying data. TAC 2011 and CrisisFACTS refutations include timing errors, where attributing sentences refer to different periods of time when facts differed, but Cyber Threat Intelligence did not. Attributing sentences for Cyber Threat Intelligence refutations were more likely to be about related, rather than directly relevant, topics for lower-ranked attributions. CrisisFACTS refutations were of a variety of types and included issues with ambiguity and parsing.
\item Parsing problems due to poorly formatted source text can lead to meaningless attributing sentences.
\item Annotators may differ on what was considered a refutation.
\end{itemize}

Details follow, starting with a new refutation typology based on an existing typology for factual errors in machine summaries from ``Understanding Factuality in Abstractive Summarization with FRANK: A Benchmark for Factuality Metrics'' by Artidoro Pangoni, Vidhisha Balachandran, and Yulia Tsetkov at NAACL 2021~\cite{pagnoni-etal-2021-understanding}. We explore TAC~2011 with NLI and sentence embedding attribution methods on human summaries for Task~1 (1 attributing sentence with preceding and following sentences for context) and Task~2 (3 attributing sentences), followed by an analysis of all three data sets with the sentence embedding attribution method on abstractive and hybrid machine summaries for Task~2.

\subsubsection{Typology of Factual Errors (Refutation)}
The typology of factual errors for abstractive summaries in~\cite{pagnoni-etal-2021-understanding} was modified for the refutation analysis as follows:
\begin{itemize}
\item Semantic Frame Errors:
  \begin{itemize}
  \item Predicate Error (PredE) : The predicate in the summary statement is inconsistent with the attribution(s).
\item Entity Error (EntE): The primary arguments (or their attributes) in the summary statement are inconsistent with the attribution(s).
\item Circumstance error: (CircE) Additional information (such as location or time) specifying the circumstance around a predicate in the summary statement is inconsistent with the attribution(s).
  \end{itemize}
\item Content Verifiability Errors
  \begin{itemize}
\item Out of Article Error (OutE): The summary statement contains information not present in the attribution(s); this includes extrinsic hallucinations.
\item Grammatical Error (GramE): The grammar mistakes in an \textit{attribution} is so wrong that it becomes meaningless.
  \end{itemize}
\item Additional Categories
  \begin{itemize}
\item Others (OthE): Refutation errors that do not correspond to any of the above categories.
\item Not an Error (NE): There are no refutation errors.
    \end{itemize}
\end{itemize}

Examples from~\cite{pagnoni-etal-2021-understanding}, modified for refutations can be found in Table~\ref{table:refutation-examples}. The first 4 (PredE, EntE, CircE, OutE) are examples of \textit{summary} statements where errors are highlighted in red. The last (GramE) is an example where the grammar in the \textit{attribution} is so wrong that it is meaningless.

\begin{table}
  \small
\centering
\begin{tabular}{llp{.3\textwidth}p{.3\textwidth}}
             & \bf{Category} & \bf{Description} & \bf{Example}\\
  \bf{PredE} & Predicate Error & The predicate in the summary statement is inconsistent with the attribution(s). & The Ebola vaccine \textcolor{red}{was rejected} by the FDA in 2019 [1].\\
  \bf{EntE} & Entity Error & The primary arguments (or their attributes) in the summary statement are inconsistent with the attribution(s). & The \textcolor{red}{COVID-19 vaccine} was approved by the FDA in 2019 [1],[3]. \\
  \bf{CircE} & Circumstance Error & Additional information (like location or time) specifying the circumstance around a predicate in the summary statement is inconsistent with the attribution(s). & The first vaccine for Ebola was approved by the FDA in \textcolor{red}{2014} [1]. \\
  \bf{OutE} & Out of Article Error & The summary statement contains information not present in the attribution(s). & \textcolor{red}{China} has started clinical trials of the COVID-19 vaccine [3].\\
  \bf{GramE} & Grammatical Error & The grammar of the \textit{attribution} is so wrong that it becomes meaningless. & The vaccine for Ebola \textcolor{red}{approved have already started}.
\end{tabular}
\caption{Examples of refutation errors, classified using the refutation typology. Source text sentences: \textit{[1]: The first vaccine for Ebola was approved by the FDA in 2019 in the US, five years after the initial outbreak in 2014. [2]:  To produce the vaccine, scientists had to sequence the DNA of Ebola, then identify possible vaccines, and finally show successful clinical trials. [3]: Scientists say a vaccine for COVID-19 is unlikely to be ready this year, although clinical trials have already started.} An additional attributing sentence which is meaningless: \textit{[4]: The vaccine for Ebola approved have already started.}} \label{table:refutation-examples}
  \end{table}

\subsection{Human: TAC 2011: Task 1 versus Task 2}
\begin{figure}[H]
  \centering
  \includegraphics[width=0.8\linewidth]{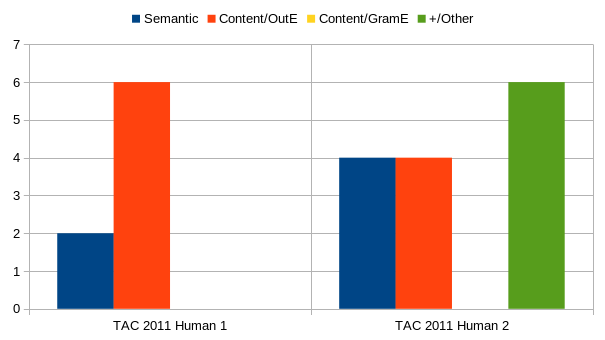}
  \caption{Refutation Error Counts for Human Summaries: TAC 2011: Task 1 and Task 2.}\label{fig:refutation-tac2011-human-bar}
\end{figure}

We evaluated refutations for the NLI and T5 sentence embedding attribution methods on human-generated summaries from TAC 2011. Two tasks were examined. In Task~1, the attributing document sentence is presented in context with the sentence preceding and following it. In Task~2, the automatic attribution methods give three supporting sentences, since some summary sentences may be the fusion of two or more sentences. Figure~\ref{fig:refutation-tac2011-human-bar} provides bar charts summarizing the refutation results of three analysts' annotations of 30 summary sentences to attribution text pairings with the NLI and embedding automatic attribution methods (60 total pairings for TAC 2011).

For Task~1, the 2 semantic refutations were time-shift errors, where the summary contained information inconsistent with an attribution because the attribution referred to a different period of time when fact(s) differed. The 8 content refutations were due to the summary containing information not found in the attribution. All refutations were for the embedding attribution method. The first annotator labeled 6 refutations (4 unique), the second labeled 3 refutations (1 unique), and the third labeled 0 (0 unique). See Table~\ref{tab:human-task-1} for more details.

The refutation profile changed for Task~2. The 4 semantic refutations were mostly but not entirely due to time-shift errors; for example, a summary and the highest ranked attributing sentence stated that the ``soles of shoes'' are considered the ultimate insult in Arab countries, while lower-ranked attributing sentences stated that ``throwing a shoe at someone'' or ``hitting someone with a shoe'' was the worst insult. The 4 content refutations were due to the summary containing additional information not found in the attributions (3) or bad attribution grammar (1). The 6 additional refutations were caused by lower-ranked attributing sentences not being directly relevant to the summary, as those attributions referenced related or unrelated topics instead. There were 5 NLI and 8 embedding refutations. The first annotator labeled 2 refutations (1 unique), the second labeled 12 refutations (9 unique) and the third labeled 0 (0 unique). It was unclear why one example was marked as a refutation and that example was not included in the counts. See Table~\ref{tab:tac2011-human-task2} for more details.

Combining the attribution and refutation analysis for TAC 2011 human summaries (Task~1 and Task~2), it appears that allowing 3 attributing sentences provided more accurate attribution but potentially add sentences of lower quality.

(Note that counting differs for the attribution pie charts versus the refutation bar charts: a summary sentence with its associated attributing sentence(s) are enumerated in the pie charts as often as the annotators mark it for that category, while counts have been deduplicated within a category for the typography-based bar charts.)

\subsection{Machine (Task~2): TAC 2011, Cyber, CrisisFACTS}
\begin{figure}[H]
  \centering
  \includegraphics[width=0.8\linewidth]{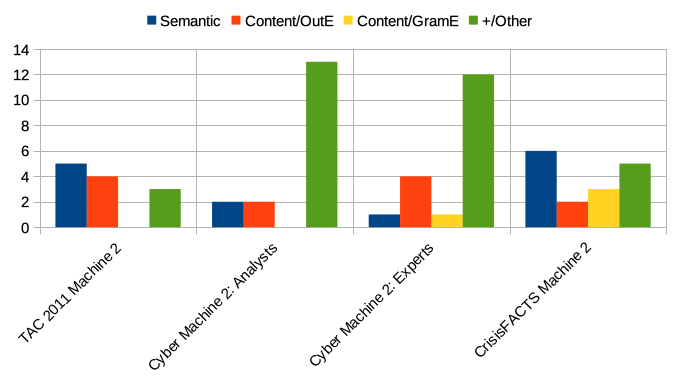}
  \caption{Refutation Error Counts for Machine Summaries: TAC 2011, Cyber Threat Intelligence, CrisisFACTS: Task 2.}\label{fig:refutation-tac2011-cti-crisisfacts-machine-bar}
\end{figure}

Next, we compared summaries generated by GPT with a hybrid approach where we asked GPT to paraphrase \texttt{occams} extracts. To limit the amount of annotation, we only evaluated the embedding attribution method and Task~2 approach of providing the top three attributing sentences. Figure~\ref{fig:refutation-tac2011-cti-crisisfacts-machine-bar} provides bar charts summarizing the refutation results of three analysts' annotations of 30 attribution text pairings for the TAC 2011, Cyber Threat Intelligence, and CrisisFACTS data sets with the abstractive and hybrid machine summarization methods (60 total pairings per data set).

TAC 2011 is a multi-document data set of fluent newswire articles over time, clustered by topic. There were 5 semantic, 4 content, and 3 additional/other refutations. The semantic refutations were due to time-shift errors, where the summary contained information inconsistent with an attributing sentence because the sentence referred to a different period of time when fact(s) differed. Two content refutations were due to summary hallucinations. In the first hallucination, the attributing sentence stated that a widow had been married to an alleged gunman for almost 10 years while the hybrid summary stated that the widow knew him for almost 10 years. In the second hallucination, 1 of 4 alleged terrorist attack masterminds were described in the abstractive summary as ``the'' mastermind. The remaining two content refutations were caused by a summary containing information not found in the attributions. The additional/other refutations were due to attributions containing related information.

There were 8 abstractive and 3 hybrid summary statements with attribution pairings that contained refutations. The first annotator labeled 5 refutations (4 unique), the second labeled 3 (1 unique), and the third labeled 8 (4 unique); it was unclear why one example was marked as a refutation and that example was not included in the counts. See Table~\ref{tab:tac2011-machine-task2} for more details.

Cyber Threat Intelligence is a single document, specialist data set. The annotators were divided into two groups: ``analysts'' for annotators with general knowledge and ``experts'' for annotators who were experts in cyber security. There were 2 semantic, 2 content, and 13 additional/other refutations in the analyst annotations, and 1 semantic, 5 content, and 12 additional/other refutations in the expert annotations. The Cyber data set had the highest proportion of additional/other refutations due to attributions containing related information. This data set contained complex technical sentences, and a summary sentence was more likely to summarize a smaller number of sentences from the source text. 

There were no semantic time-shift errors: while a document may describe events over a time period, there was no additional complication from events split over different documents. Both groups identified a semantic refutation for a hybrid summary in which the LLM paraphrasing used stronger language (``highly likely'') compared to the source text (``likely'') to describe the probability of network compromise. The experts identified two summary hallucinations, while the analysts only identified one of the two summary hallucinations.


There were 10 abstractive and 6 hybrid summary statements with attribution pairings that contained refutations in the analyst annotations. The first annotator labeled 16 refutations (13 unique), the second labeled 3 (0 unique), and the third labeled 0 (0 unique); it was unclear why one example was marked as a refutation and it is not included in the counts. There were 9 abstractive and 8 hybrid summary statements with attribution pairings that contained refutations in the expert annotations; 7 of those refutations were also marked as refutations by the analysts. The first annotator labeled 1 refutation, the second labeled 2, and the third labeled 14 refutations (including the two summary hallucinations). See Table~\ref{tab:cyber-machine-task2-analyst} and Table~\ref{tab:cyber-machine-task2-expert} for more details.

CrisisFACTS is a multi-stream data set and is closest to raw data of the three data sets. There were 6 semantic, 5 content, and 5 additional/other refutations. Content refutations were divided between 2 OutE refutations (the summary contained information not present in the attributions) and 3 GramE refutations (the grammar of an attributing sentence was so wrong that it became meaningless). TAC 2011 contains fluent newswire and was straightforward to parse into sentences for attribution. Cyber Threat Intelligence documents contained fluent text, lists of fluent text, and unformatted tables. An unformatted table was parsed as a very long sentence, and additional computing resources were required for the automatic embedding attribution method to accommodate very long sentences. Although cyber table-based sentences were not meaningful without formatting, these sentences did not appear in human evaluation. CrisisFACTS contained short text excerpts from multiple data streams (Twitter, Reddit, newswire). Text excerpts were not necessarily fluent text and might be truncated either in the original data set or afterwards in post-processing when multiple excerpts were concatenated and possibly truncated to fit within the LLM context window. CrisisFACTS was a challenging data set to automatically parse into ``sentences'' for attribution, and this was reflected in refutations where an attributing ``sentence'' was meaningless due to poor parsing and truncation.

The 6 CrisisFACTS semantic refutations were mostly due to time-shift errors leading to inconsistent information, with an additional semantic error due to additional location information in the summary being inconsistent with one of the attributing sentences. The additional/other refutations were due to attributions containing related information or the summary statement being so ambiguous that it was difficult to determine if the paired attributing sentences were relevant to the underlying disaster.

There were 7 abstractive and 5 hybrid summary statements with attribution pairings that contained refutations. The first annotator labeled 11 refutations (8 unique) the second labeled 4 (1 unique), and the third labeled 0 (0 unique); it was unclear why one example was marked as a refutation, and it is not included in the counts. See Table~\ref{tab:crisisfacts-machine-task2} for more details.

\section{Conclusions}

In this paper, we set up a series of experiments to measure how well automatic attribution methods worked for attributing information generated by a large language model. We first compared two automatic approaches, one based on a natural language inference model, trained on entailment data, and compared it with a baseline using a sentence embedding method. We found that the embedding model, which was the larger unsupervised model, provided better attribution. However, it had higher instances of refutations. 

Next, we compared the embedding approach on two styles of summaries on three datasets: TAC 2011, a Cyber Threat Intelligence dataset, and the TREC 2022 CrisisFACTS dataset. Here, we found that the hybrid summary, a paraphrase of an extractive \texttt{occams} summary, generated text whose sentences were easier to attribute to information in the original documents for all three datasets.

Finally, we used a custom topology to identify the proportion of different categories of attribution-related errors. These refutations often differed from hallucinations in machine summaries. Proportions varied based on the underlying data, and parsing issues could lead to meaningless attributing sentences due to poorly formatted source text.
\vspace{6pt} 

\section*{Acknowledgements}
The authors thank Ani Nenkova for her visit to the Summer Conference on Applied Data Science in 2023 and for her suggestions, which help initiate this work. We like to thank a colleague, who chose to remain anonymous for their critical contribution to this work. We also are very grateful to the analyst annotators and cyber experts for their evaluation of automatic attribution methods for three datasets.

\appendixtitles{yes} 
\appendixstart
\appendix
\section[\appendixname~\thesection]{Annotation Guidelines}
This appendix show sample annotation guidelines for TAC 2011. We ask annotators to read these guidelines as part of the human evaluation. The SCADS Cyber Threat Intelligence Data Set and CrisisFACTS guidelines were similar (3 attributing sentences for Task 2: Machine Summaries).

\subsection[\appendixname~\thesubsection]{TAC 2011: Task 1: Human Summaries}
\begin{myquotation}

\newcommand{\mysample}[1]{
\begin{center}
\begin{minipage}{.80\textwidth}\fontfamily{lmss}\selectfont
#1
\end{minipage}
\end{center}
}


\maketitle

\section*{Overview}
We are a team interested in evaluating attribution for summarization (how to link automatically generated summaries back to their source material.)

In this project, you will evaluate the quality of attributions for summaries that have been created by humans or by automatic methods that mimic what humans would write. We would like to evaluate if an attribution method can provide accurate attributions, and if the accuracy is influenced by the type of automatic summarization method used.

At a high level, the project breaks down into two tasks:

\begin{enumerate}
    \item Given a statement in a summary, judge the quality of a single attribution.
    \item Given a statement in a summary, judge whether the statement is fully supported by three (ranked) attributions.
\end{enumerate}

Examples of the task you have chosen will be provided in this document.

 This evaluation will use the TAC 2011 (Text Analysis Conference) data set for multi-document summarization. The data set contains news articles grouped by topic (10 documents per topic). A summary is based on a single topic (10 documents).

\section*{Task 1: Judge the quality of a single attribution}
In this part, you will be provided a statement from the summary, and will be asked to judge the quality of a corresponding attribution in the source material. The statement will be a single sentence from the summary. The attribution will be a single (1) sentence from the source document, along with surrounding context sentences (sentence before and sentence after the attribution sentence in the source document, if available).

Carefully read the summary statement and the attribution sentence with the surrounding context. The summary statement may appear to be very fluent and well-formed, and there may be slight differences compared to the corresponding attribution in the source document that is not easy to discern at first glance. Please pay close attention to the text, as if you were proofreading.

Your task will be to determine which category applies:
\begin{itemize}
\item \textbf{Support}: Some or all of the information in the summary statement is supported by the attribution sentence.
\item \textbf{No Support}: The attribution sentence does not support the summary statement. 
\item \textbf{Unclear (Can’t Make Judgement)}: The information provided does not make it possible to determine whether the attribution sentence supports or does not support the summary statement. The information is unclear, incomprehensible, malformed, or deficient in some other way.
\end{itemize}

The second part of the task will be to determine if the answer is ``yes'' or ``no'' to the following question:
\begin{itemize}
\item \textbf{Refute}: Does the attribution contain information that refutes the summary statement? There can be a refutation (``yes'') with ``support'' or ``no support''. Answer ``no'' if the information is unclear.
\end{itemize}
\subsection*{Examples}

For TAC 2011, a summary is based on 10 documents for a single topic. The upcoming examples are from a single topic on threats to sea turtles.

\subsubsection*{Example: Support, Refute=no}
\mysample{
[SUMMARY] Sea turtles are threatened by increased poaching, habitat destruction for land development, illegal fishing, and pollution.\newline

[CONTEXT] The fishermen do not properly release the turtles and often kill them, leaving them to wash ashore. \newline
[**EVAL**] Turtles that come to lay eggs on the beaches may be killed by pollution, stray dogs or foxes, or captured by tourists. \newline
[CONTEXT] Mohammad Aminul Islam, the top administrator for the area, ordered local officials to teach people, from fishermen to tourists, to change their behavior.
}

The sentence to evaluate for attribution is “Turtles that come to lay eggs on the beaches may be killed by pollution, stray dogs or foxes, or captured by tourists.” This sentence provides supporting evidence that sea turtles are threatened by pollution. While the first context sentence supports illegal fishing, the decision for the evaluation should only be based on the sentence with the {\fontfamily{lmss}\selectfont[**EVAL**]} (evaluate) prefix.

The summary statement is not fully supported since no supporting evidence is provided for poaching, habitat destruction for land development, and illegal fishing by the attribution sentence. However, partial support meets the definition of support. You may mark “support” for an answer.

Consider the refute question next. The attribution sentence supports the summary statement and also contains supplemental information. It does not contain contradictory information. The answer is ``refute=no''.

\subsubsection*{Example: No Support, Refute=yes}
\mysample{
[SUMMARY] Malaysia has arrested a number of poachers, many from south China, for poaching turtles and other animals.\newline

[CONTEXT] At that price, a single nest containing about 100 eggs nets a poacher 1,000 pesos. Compare that, he says, to Mexico's minimum daily wage of 45 pesos. \newline
[**EVAL**] The Mexican government takes poaching seriously, he continued, assessing steep fines or jail time--``it's a worse offense than drugs''--but it lacks the manpower to prevent the crime in the first place or prosecute offenders in the second. \newline
[CONTEXT] Rapid development endangers the turtles. 
}

The attribution sentence is the second of three consecutive sentences. This sentence does not provide supporting evidence for poaching in Malaysia because the attribution is about poaching in Mexico which is a different country. (The difference is due to the summary being based on a different news article than the attribution.) You may mark ``no support.''

Consider the refute question next. While the attribution sentence supports turtle poaching having legal repercussions, this example will be marked as “refute=yes” because the country differs. 

This example could have also been marked as “refute=no” because the statement and attribution sentence are about different countries. There will be ambiguous cases like this where it is not clear if an attribution refutes a statement or not. We want you to use your best judgment during the evaluation. 

If it helps to decide between “refute=yes” and “refute=no”, the purpose of the “refute” choice is to identify more focused algorithmic misfires. For example, when an automatic summarization method generates an incorrect fact (“hallucination”) like substituting one nationality for another, or misinterprets the source document during text generation (e.g., dog bites man misinterpreted as man bites dog). Or when an attribution method provides a link which is related to the summary statement but misses the mark (e.g., Mexican poaching instead of Malaysian poaching).

Another type of misfire is when the attribution method has low accuracy and picks an irrelevant sentence. We would like to capture this with the “refute=no”.  

\subsubsection*{Example: No Support, Refute=no}
\mysample{
[SUMMARY] 90 sea turtles were rescued in Mexico when they became comatose from a cold snap.\newline

[CONTEXT] Some sea turtle species eat jellyfish like they are going out of style.\newline 
[**EVAL**] If the turtles weren't around to eat the jellyfish, the jellyfish population would explode.\newline 
[CONTEXT] The hungry jellyfish hordes would gorge on zooplankton.
}

The attribution sentence is “If the turtles weren't around to eat the jellyfish, the jellyfish population would explode.” This sentence does not provide supporting evidence for a turtle rescue. You may mark ``no support.''

 Consider the refute question. There will be a topic-based relationship between a statement and an attribution with TAC 2011 since the summary is based on news articles from the same topic (e.g., threats to sea turtles). However, the connection is too tenuous for this example, and it should be marked “refute=no”.

\section*{Questions or Feedback?}
If you have questions about the task, or any feedback about how we could make it better or what your experience was like with it, please contact us on Slack using the channel \texttt{\#inf-attribution-evaluation} and we'll get back to you. Thanks!

 
\end{myquotation}

\subsection[\appendixname~\thesubsection]{TAC 2011: Task 2: Machine Summaries}

\begin{quotation}

\newcommand{\mysample}[1]{
\begin{center}
\begin{minipage}{.80\textwidth}\fontfamily{lmss}\selectfont
#1
\end{minipage}
\end{center}
}




\section*{Overview}
We are a team interested in evaluating attribution for summarization (how to link automatically generated summaries back to their source material.)

In this project, you will evaluate the quality of attributions for summaries that have been created by humans or by automatic methods that mimic what humans would write. We would like to evaluate if an attribution method can provide accurate attributions, and if the accuracy is influenced by the type of automatic summarization method used.

At a high level, the project breaks down into two tasks:

\begin{enumerate}
    \item Given a statement in a summary, judge the quality of a single attribution.
    \item Given a statement in a summary, judge whether the statement is fully supported by three (ranked) attributions.
\end{enumerate}

Examples of the task you have chosen will be provided in this document.

 This evaluation will use the TAC 2011 (Text Analysis Conference) data set for multi-document summarization. The data set contains news articles grouped by topic (10 documents per topic). A summary is based on a single topic (10 documents).

\section*{Task 2: Judge the quality of a group of attributions}

In this part, you will be provided a statement from the summary, and will be asked to judge the amount of support by the attribution. The statement will be a single sentence from the summary. The attribution will be three (3) sentences from the source material.

Carefully read the summary statement and the 3 attribution sentences. The summary statement may appear to be very fluent and well-formed, and there may be slight differences compared to the corresponding 3 attribution sentences from the source data that is not easy to discern at first glance. Please pay close attention to the text, as if you were proofreading.

Your task will be to determine which category applies:
\begin{itemize}
\item \textbf{Full Support}: All of the information in the summary statement is supported by the attribution sentences.
\item \textbf{Partial Support}: Only some of the information in the summary statement is supported, but other parts of the information are missing.
\item \textbf{No Support}: The attribution sentences do not support the summary statement.  
\item \textbf{Unclear (Can't Make Judgment)}: The information provided does not make it possible to determine whether the attribution sentence fully supports, partially supports, or does not support the summary statement. The information is unclear, incomprehensible, malformed, or deficient in some other way. 
\end{itemize}

The second part of the task will be to determine if the answer is “yes” or “no” to the following question:
\begin{itemize}
\item \textbf{Refute}: Does the attribution contain information that refutes the summary statement? There can be a refutation (“yes”) with “full support”, “partial support”, or “no support”. Answer “no” if the information is unclear.
\end{itemize}
\newcounter{question}
\setcounter{question}{0}

\newcommand{\question}[1]{\item[Q\refstepcounter{question}\thequestion.] #1}
\newcommand{\answer}[1]{\item[A\thequestion.] #1}

\subsubsection*{FAQ}
\begin{itemize}
\question\textit{How do we mark a group of attribution sentences if one of the sentences fully supports the summary statement, no matter what the rest say? Do we mark this as ``full support'' or ``partial support''?}
\answer If one attribution sentence fully supports the summary statement, then mark the group as ``full support'' for the first question.

\question\textit{In InfinityPool (the annotator platform), choices were not pre-selected for the first question about support. However, an answer was already highlighted for the second refute question before I answered it. Should this happen?}
\answer The refute questions all default to ``no''. Change an answer to ``yes'' (when applicable).
\end{itemize}

\subsection*{Examples}

For TAC 2011, a summary is based on 10 documents for a single topic. The upcoming examples are from a single topic on threats to sea turtles.

\subsubsection*{Example: Full support, Refute=no}
\mysample{
[SUMMARY] Some countries have passed laws to protect the turtles.\newline

[**EVAL**] The Mexican government takes poaching seriously, he continued, assessing steep fines or jail time--``it's a worse offense than drugs''--but it lacks the manpower to prevent the crime in the first place or prosecute offenders in the second.\newline
[**EVAL**] A Malaysian court Friday jailed a group of 13 Chinese fishermen for 18 months each for poaching nearly 80 protected turtles.\newline
[**EVAL**] Ahmed would not give any specific reason for the spike in deaths, but said the use of illegal fishing nets near the shoreline has apparently increased recently.
}

The first sentence describes a poaching law in Mexico. Malaysian law can be inferred by the second sentence. The third sentence describes illegal fishing nets but does not specify the country (Bangladesh). Combined, the sentences are sufficient to fully support a statement that more than one country has laws to protect turtles, and the example can be marked ``full support''.

Consider the refute question next. Each sentence supports the summary statement and does not contain contradictory information. The answer is ``refute=no''.

\subsubsection*{Full support, Refute=no}
\mysample{
[SUMMARY] 90 sea turtles were rescued in Mexico when they became comatose from a cold snap.\newline

[**EVAL**] Nearly 90 sea turtles rescued after a cold snap left them comatose have been returned to the Gulf of Mexico.\newline
[**EVAL**] Dozens of rehabilitated sea turtles, threatened by cold, returned to Gulf of Mexico.\newline
[**EVAL**] Green turtles are born in southern Mexico and spend their early years feeding on turtle grass in shallow areas such as the Laguna Madre.
}

The first sentence fully supports the summary statement. The second sentence partially supports the statement. The third sentence is about turtles in Mexico but not a rescue. 
If one or two sentences fully support the summary statement, the question may be marked ``full support''. Support is not required  from all 3 sentences.

Consider the refute question next. The first two sentences support the summary sentence and do not contain contradictory information. The third sentence is about Mexican turtles but is about a different topic (rescue versus early life). The connection is tenuous and will not be viewed as a refutation. Since none of the sentences contain a refutation, the answer to the question is “refute=no”.

This example could have also been marked as “refute=yes” because the third sentence is about Mexican turtles so there is a connection between that sentence and the summary statement, but the facts are different. There will be ambiguous cases like this where it is not clear if an attribution sentence refutes a statement or not. We want you to use your best judgment during the evaluation.

If it helps to decide if a sentence refutes the summary statement or not, the purpose of the “refute” choice is to identify more focused algorithmic misfires. For example, when an automatic summarization method generates an incorrect fact (“hallucination”) like substituting one nationality over another, or misinterprets the source document during text generation (e.g., dog bites man misinterpreted as man bites dog). Or when an attribution method provides a link which is related to the summary statement but misses the mark.  

Another type of misfire is when the attribution method has low accuracy and picks an irrelevant sentence. We would like to capture this with ``refute=no''.

\subsubsection*{Full support, Refute=yes}
\mysample{
[SUMMARY] 17 Chinese nationals found with 274 protected turtles, most of them dead, were recently arrested in Malaysian waters.\newline

[**EVAL**] The poachers were caught by Malaysian marine police on March 28 in waters off the coast of eastern Sabah state with a catch of 274 protected Hawksbill and Green turtles, only 32 of which were still alive.\newline
[**EVAL**] They were the second group of fishermen from China's Hainan island jailed this week for turtle poaching off the waters of Malaysia's eastern Sabah state, with 17 men sentenced on Wednesday for catching 274 of the creatures.\newline
[**EVAL**] A Malaysian court Friday jailed a group of 13 Chinese fishermen for 18 months each for poaching nearly 80 protected turtles. 
}

The first sentence provides supporting evidence for poachers being caught with 274 protected turtles, most of them dead. The second sentence provides support for the arrest of 17 Chinese nationals with 274 turtles. The third sentence is about a different arrest.
The first two sentences provide full support for the summary statement, and this example may be marked ``full support''.

Consider the refute question next. The third sentence is on the same subject: a Malaysian arrest of Chinese nationals for poaching protected turtles. However, it is about a different arrest. This is an example of a misfire where a sentence is related but misses the mark, and can be viewed as an instance of ``refute=yes''.

\subsubsection*{Partial support, Refute=no}
\mysample{
[SUMMARY] Sea turtles are threatened by increased poaching, habitat destruction for land development, illegal fishing, and pollution.\newline

[**EVAL**]  The turtles--which can be found in other parts of Asia along the Andaman Coast and the South China Sea--have seen their numbers reduced drastically in recent years mostly due to poaching of their eggs, pollution and habitat loss.\newline
[**EVAL**] Turtles that come to lay eggs on the beaches may be killed by pollution, stray dogs or foxes, or captured by tourists.\newline
[**EVAL**] Sea turtles face a host of threats around the world. 
}

The three sentences provide supporting evidence for a threat to sea turtles for several reasons but do not support illegal fishing. This example may be marked “partial support”.

None of the sentences refute the summary sentence. The answer is “refute=no”.
 
\subsubsection*{Example: No support, Refute=no}
\mysample{
[SUMMARY] Scientists say that the loss of sea turtles could have unforeseen, even catastrophic, consequences.\newline

[**EVAL**] Efforts in Mexico to protect sea turtles are a patchwork of official and unofficial endeavors, according to Sea Turtle, Inc., a nonprofit based in South Padre Island.\newline
[**EVAL**] ``We are trying to mobilize resources to make a bigger plan to save the sea turtles in the future.''\newline
[**EVAL**] Thai villagers have caught a river terrapin turtle that was thought to be extinct in the country.
}

Sentences will be related to a common topic (threats to sea turtles). However, the 3 attribution sentences do not support the summary statement. This example may be marked “no support”. 

Consider the refute question next. Each attribution sentence is irrelevant to the summary statement. An irrelevant sentence is not a refutation. None of the sentences refute the summary statement and the question may be marked “refute=no”.

\section*{Questions or Feedback?}
If you have questions about the task, or any feedback about how we could make it better or what your experience was like with it, please contact us on Slack using the channel \texttt{\#inf-attribution-evaluation} and we'll get back to you. Thanks!


\end{quotation}



\section[\appendixname~\thesection]{Refutation Analysis Tables}

Tables used for the refutation analysis in Subsection~\ref{sec:refutation-analysis} are provided in this appendix. Duplicates are shown in a gray font.

\begin{table}[p]\small
  \parbox[b]{4.85in}{}
  \centering
  \begin{tabular}{|l|c|c|c|p{1.5cm}|p{1cm}|p{4cm}|}
    \hline
    \textbf{ID} & \textbf{Method} & \textbf{Annotator} & \textbf{L} &
    \textbf{Primary} & \textbf{2nd} & \textbf{Comment} \\
    \hline
    13135 & Emb & 8d4ff & S &  Content  & OutE & Primary part of summary not present in attribution (=13245)\\ \hline
    13181 & Emb & 8d4ff & N & Content & OutE & Summary contains information not present in attribution which is about a related topic \\ \hline
    13182 & Emb & 8f14c & U & Semantic & EntE & Time Shift led to inconsistent info (=13221)  \\ \hline
   13189 & Emb & 8d4ff & S & Semantic Content & CircE OutE & Time Shift led to inconsistent timing info; Summary contains info not in attributing sentence but in context \\ \hline
   13190 & Emb & 8d4ff & S & Content & OutE & Timing information in summary not present in attribution \\ \hline
   \gray{13221} & \gray{Emb} & \gray{8d4ff} & \gray{S} & \gray{Semantic} & \gray{EntE} & \gray{Time Shift led to inconsistent info (=13182)} \\ \hline   
   13229 & Emb & 8d4ff & S & Content & OutE & Timing info stated differently (August versus Thursday) \\ \hline
   \gray{13245} & \gray{Emb} & \gray{8f14c} & \gray{U} & \gray{Content}  & \gray{OutE} & \gray{Primary part of summary not present in attribution (=13135)}\\ \hline
   13258 & Emb & 8f14c & N & Content & OutE & Timing and supplemental location information in summary not present in attribution \\ \hline
  \end{tabular}
  \parbox[b]{4.85in}{}
  \caption{TAC 2011: Human Summaries: Task 1: One Attributing Sentence with Context. L/Label: S=Support, N=No Support, U=Unclear.}
  \label{tab:human-task-1}
\end{table}

\begin{table}[h]\small
  \parbox[b]{4.85in}{}
  \centering
  \begin{tabular}{|l|c|c|c|p{1.6cm}|p{1cm}|p{4.1cm}|}
    \hline
    \textbf{ID} & \textbf{Method} & \textbf{Annotator} & \textbf{L} &
    \textbf{Primary} & \textbf{2nd} & \textbf{Comment} \\
    \hline
    12981 & Emb & eed54 & P & Semantic & CircE & Time Shift error led to inconsistent timing info (2,3) \\ \hline
    13022 & Emb & 415f6 & P & Content & OutE &  Summary contains statistic not present in attributions\\ \hline
    13025 & Emb & 415f6 & F & Content Additional & GramE OthE & Bad attribution grammar (2); Attribution on related topic (3)\\ \hline
    13039 & NLI & eed54 & F & Semantic & PredE  & Time Shift led to inconsistent info (3) (=13144)\\ \hline
    13041 & NLI & 415f6 & F & Semantic & EntE & Attributions inconsistent but related (2,3) (=13046)\\ \hline
    \gray{13046} & \gray{Emb} & \gray{415f6} & \gray{F} & \gray{Semantic} & \gray{EntE} & \gray{Attributions inconsistent but related (2,3) (=13041)}\\ \hline
    13056 & NLI & 415f6 & F & Additional & OthE & Unrelated attribution (3)\\ \hline
    13065 & Emb & 415f6 & F & Content & OutE &  Summary statement contains additional timing information not present in attributions\\ \hline
    13072 & Emb & 415f6 & F & Semantic & PredE & Inconsistent info, from time shift or different disaster (3)\\ \hline
    13095 & Emb & 415f6 & P & Content Additional & OutE OthE & Summary contains info not present in attributions; Attribution for wrong location (3)\\ \hline
    13097 & NLI & 415f6 & F & Additional & OthE  & Attribution unrelated (3)\\ \hline
    13098 & Emb & 415f6 & F & Additional & NE & Unclear why selected as refutation\\ \hline
    13104 & NLI & 415f6 & F & Additional & OthE & Attributions related (2,3) \\ \hline
    13105 & Emb & 415f6 & F & Additional & OthE & Attributions related (2,3) \\ \hline
    \gray{13144} & \gray{NLI} & \gray{415f6} & \gray{P} & \gray{Semantic} & \gray{PredE} & \gray{Time Shift led to inconsistent info (3) (=13039)}\\ \hline
  \end{tabular}
  \parbox[b]{4.85in}{}
  \caption{TAC 2011: Human Summaries: Task 2: Three Attributing Sentences. L/Label: F=Full Support, P=Partial Support.}
  \label{tab:tac2011-human-task2}
\end{table}

\begin{table}[h]\small
  \centering
  \begin{tabular}{|l|c|c|c|p{1.6cm}|p{1.1cm}|p{4.1cm}|}
    \hline
    \textbf{ID} & \textbf{Method} & \textbf{Annot.} & \textbf{L} &
    \textbf{Primary} & \textbf{2nd} & \textbf{Comment} \\
    \hline
    13951 & Abstractive & 8d4ff & P & Content & OutE & Summary contains information not present in attributions but might be inferred\\ \hline
    13954 & Abstractive & 8d4ff & P & Semantic Content & CircE OutE & Time shift (1,2); Summary contained info not present in attributions\\ \hline
    13957 & Abstractive & 8d4ff & P & Semantic & PredE & Time shift led to inconsistent info (2,3) \\ \hline 
    13961 & Abstractive & 8d4ff & F & Semantic & PredE & Time shift led to inconsistent info (3)\\ \hline
    13971 & Hybrid & 8d4ff & F & Content & OutE & Summary hallucination (1) (=14091)\\ \hline
    \gray{14021} & \gray{Abstractive} & \gray{8f14c} & \gray{P} & \gray{Semantic} & \gray{PredE} & \gray{Time shift led to inconsistent info (3) (=13961, 14081)}\\ \hline
    14026 & Abstractive & 8f14c & F & Additional & OthE & Attribution contains related info (2)\\ \hline
    14045 & Hybrid & 8f14c & F & Semantic & PredE & Time shift led to inconsistent info (2) (=14105) \\ \hline
    \gray{14077} & \gray{Abstractive} & \gray{36f2c} & \gray{P} & \gray{Semantic} & \gray{PredE} & \gray{Time shift led to inconsistent info (2,3) (=13597)}\\ \hline
    \gray{14081} & \gray{Abstractive} & \gray{36f2c} & \gray{F} & \gray{Semantic} & \gray{PredE} & \gray{Time shift led to inconsistent info (3) (=13961,14021)}\\ \hline
    \gray{14091} & \gray{Hybrid} & \gray{36f2c} & \gray{P} & \gray{Content} & \gray{OutE} & \gray{Summary hallucination (1) (=13971)}\\ \hline
    14092 & Abstractive & 36f2c & F & Additional & OthE & Attribution same topic but different location (2) \\ \hline
    14049 & Abstractive & 36f2c & F & Semantic & PredE & Time shift led to inconsistent info (3)\\ \hline
    14102 & Abstractive & 36f2c & F & Content & OutE & Summary hallucination (summary used ``the'' versus 1 of 4)\\ \hline
    \gray{14105} & \gray{Hybrid} & \gray{36f2c} & \gray{F} & \gray{Semantic} & \gray{PredE} & \gray{Time shift led to inconsistent info (2) (=14045)}\\ \hline
    14108 & Abstractive & 36f2c & F & Additional & NE & Unclear why selected as refutation \\ \hline
    14115 & Hybrid  & 36f2c & F & Additional & OthE & Attributions contain related information (2,3)\\ \hline
  \end{tabular}
  \caption{TAC 2011: Machine Summaries: Task 2: 3 Attributing Sentences. L/Label: F=Full Support, P=Partial Support.}
  \label{tab:tac2011-machine-task2}
\end{table}

\begin{table}[h] \small
  \centering
  \begin{tabular}{|l|c|c|c|p{1.6cm}|p{1.1cm}|p{4.1cm}|}
    \hline
    \textbf{ID} & \textbf{Method} & \textbf{L} & \textbf{Annot.} &
    \textbf{Primary} & \textbf{2nd} & \textbf{Comment} \\
    \hline
    17040 & Abstractive & P & 73655 & Content  & OutE & Summary hallucination (=17160)\\ \hline
    17041 & Abstractive & P & 73655 & Additional & OthE & Related information (2,3), Summary $\approx$ Attribution~1 (=17161)\\ \hline
    17043 & Hybrid & P & 73655 & Additional & OthE & Related information (2,3), Summary $\approx$ Attribution~1\\ \hline
    17044 & Hybrid & F & 73655 & Additional & OthE & Closely related information (2), related information (3), Summary $\approx$ Attribution~1\\ \hline
    17047 & Abstractive & P & 73655 & Additional & OthE & Related information (2)\\ \hline
    17048 & Abstractive & F & 73655 & Additional & OthE & Related information (2,3), Summary $\approx$ Attribution~1 \\ \hline
    17049 & Hybrid & P & 73655 & Semantic Additional & PredE OthE & Summary close to Attribution~1 but uses stronger language (``very likely'' versus ``likely''); Related information (2,3)\\ \hline
    17053 & Abstractive & F & 73655 & Additional & OthE & Related information (3)\\ \hline
    17056 & Hybrid & P & 73655 & Additional & OthE & Related information (3)\\ \hline
    17057 & Abstractive & P & 73655 & Content & OutE & Summary contains info not present in attributions\\ \hline
    17061 & Abstractive & P & 73655 & Additional & OthE & Related information (3) (=17181)\\ \hline
    17063 & Hybrid & F & 73655 & Additional & OthE & Related information (2,3), Summary $\approx$ Attribution~1\\ \hline
    17065 & Hybrid & P & 73655 & Additional & NE & Unclear why selected as refutation\\ \hline
    17066 & Abstractive & P & 73655 & Additional & OthE & Closely related information (2,3), Summary $\approx$ Attribution~1 \\ \hline
    17072 & Hybrid & F & 73655 & Additional & OthE & Closely related information (2), related information (3), Summary $\approx$ Attribution~1\\ \hline
    17073 & Abstractive & P & 73655 & Semantic & EntE & Closely related information (2,3), Summary $\approx$ Attribution~1 but summary entity lacks sufficient information (``incidents'' versus ``incidents involving landing URLs'')\\ \hline
    17083 & Abstractive & P & 73655 & Additional & OthE & Closely related information (2,3), Summary $\approx$ Attribution~1\\ \hline
    \gray{17160} & \gray{Abstractive} & \gray{N} & \gray{8f14c} & \gray{Content} & \gray{OutE} & \gray{Summary hallucination (=17040)}\\ \hline
    \gray{17161} & \gray{Abstractive} & \gray{P} & \gray{8f14c} & \gray{Additional} & \gray{OthE} & \gray{Related information (2,3) (=17041)}\\ \hline
    \gray{17181} & \gray{Abstractive} & \gray{P} & \gray{8f14c} & \gray{Additional} & \gray{OthE} & \gray{Related information (3) (=17061)}\\ \hline
  \end{tabular}
  \caption{Cyber Threat Intelligence: Machine Summaries: Analysts: Task 2: 3 Attributing Sentences. L/Label: F=Full Support, P=Partial Support, N=No Support.}
  \label{tab:cyber-machine-task2-analyst}
\end{table}

\begin{table}[h] \small
  \centering
  \begin{tabular}{|l|c|c|c|p{1.6cm}|p{1.1cm}|p{4.1cm}|}
    \hline
    \textbf{ID} & \textbf{Method} & \textbf{L} & \textbf{Annot.} &
    \textbf{Primary} & \textbf{2ndary} & \textbf{Comment} \\
    \hline
    16212 & Abstractive & P & 415f6 & Additional & OthE & Related information (3) (=17061, 17181)\\ \hline
    17592 & Hybrid & F & 5e2f7 & Semantic Additional & PredE OthE & Summary close to Attribution~1 but uses stronger language (``very likely'' versus ``likely''); Related information (2,3) (=17049) \\ \hline
    17606 & Hybrid & F & 5e2f7 & Additional & OthE & Related information (2,3), Summary $\approx$ Attribution~1 (=17063) \\ \hline
    17633 & Abstractive & F & 84863 & Additional & NE & Unclear why selected as refutation  \\ \hline
    17639 & Abstractive & P & 84863 & Content & OutE & Summary hallucination \\ \hline
    17642 & Hybrid & F & 84863 & Additional & OthE & Related information (2,3), Summary $\approx$ Attribution 1 \\ \hline
    17643 & Abstractive & P & 84863 & Content Additional & OutE OthE & Summary contains information not in attributions, Related information (3) \\ \hline
    17645 & Abstractive & P & 84863 & Content  & OutE & Summary hallucination (=17040, 17160) \\ \hline
    17652 & Abstractive & N & 84863 & Additional & OthE & Related information (2) (=17047) \\ \hline
    17658 & Abstractive & N & 84863 & Additional & OthE & Related information (3) (=17053) \\ \hline
    17660 & Abstractive & P & 84863 & Content & OutE & Summary contains named entity while (3) uses pronoun \\ \hline
    17661 & Hybrid & P & 84863 & Additional & OthE & Related information (3) (=17056) \\ \hline
    17667 & Abstractive & F & 84863 & Content & GramE & Badly parsed attribution (2) \\ \hline
    17679 & Hybrid & F & 84863 & Additional & OthE & Related information (2,3) \\ \hline
    17680 & Hybrid & F & 84863 & Additional & OthE & Related information (2,3), Summary $\approx$ Attribution~1 \\ \hline
    17684 & Hybrid & F & 84863 & Additional & OthE & Related information (2,3), Summary $\approx$ Attribution~1 \\ \hline
    17687 & Hybrid & F & 84863 & Additional & OthE & Related information (3) \\ \hline
  \end{tabular}
  \caption{Cyber Threat Intelligence: Machine Summaries: Experts: Task 2: 3 Attributing Sentences. L/Label: F=Full Support, P=Partial Support, N=No Support.}
  \label{tab:cyber-machine-task2-expert}
\end{table}

\begin{table}[h] \small
  \centering
  \begin{tabular}{|l|c|c|c|p{1.6cm}|p{1.1cm}|p{4.1cm}|}
    \hline
    \textbf{ID} & \textbf{Method} & \textbf{L} & \textbf{Annot.} &
    \textbf{Primary} & \textbf{2nd} & \textbf{Comment} \\
    \hline
    17341 & Abstractive & P & 73655 & Content Additional & GramE OthE & Badly parsed, related information (2,3); Not enough information in summary to identify which fire\\ \hline
    17344 & Abstractive & U & 73655 & Additional & OthE & Not enough information in summary to identify which disaster\\ \hline
    17347 & Hybrid & F & 73655 & Semantic Additional & PredE OthE & Time shift led to inconsistent info (3); Attribution~2 for a different fire (=17467)\\ \hline
    17420 & Hybrid & P & 73555 & Semantic & PredE  & Time shift led to inconsistent info (3)\\ \hline
    17425 & Abstractive & U & 73655 & Semantic Content & CircE GramE & Additional location information in summary inconsistent with attribution (2); Badly parsed attributions difficult to read (2,3)\\ \hline
    17428 & Hybrid & F & 73655 & Semantic & PredE & Time shift led to inconsistent info (3), Summary $\approx$ Attribution~1 (=17488)\\ \hline
    17429 & Hybrid & P & 73655 & Semantic & PredE & Time shift led to inconsistent info (3), Summary $\approx$ Attribution~1 (=17489)\\ \hline
    17432 & Abstractive & P & 73655 & Content & OutE & Summary contains info not present in attributions\\ \hline
    17436 & Abstractive & N & 73655 & Additional & OthE & Attribution provided option not listed in summary (3)\\ \hline
    17439 & Hybrid & U & 73655 & Content Additional & OutE OthE  & Summary contains world knowledge not included in attribution (1); Attributions on related info (2,3)\\ \hline
    17448 & Abstractive & F & 73655 & Semantic & PredE & Time shift error led to inconsistent info (1,3)\\ \hline
    17452 & Abstractive & F & 73655 & Additional & NE & Unclear why selected as refutation \\ \hline
    \gray{17467} & \gray{Hybrid} & \gray{F} & \gray{36f2c} & \gray{Semantic Additional} & \gray{PredE OthE} & \gray{Time shift led to inconsistent info (3); Attribution~2 for a different fire (=17347)}\\ \hline
    \gray{17488} & \gray{Hybrid} & \gray{F} & \gray{36f2c} & \gray{Semantic} & \gray{PredE} & \gray{Time shift led to inconsistent info (3) (=17428)}\\ \hline
    \gray{17489} & \gray{Hybrid} & \gray{P} & \gray{36f2c} & \gray{Semantic} & \gray{PredE} & \gray{Time shift led to inconsistent info (3), Summary $\approx$ Attribution~1 (=17429)}\\ \hline
    17515 & Abstractive & F & 36f2c & Content & GramE & Badly parsed attribution (3)\\ \hline
  \end{tabular}
  \caption{CrisisFACTS: Machine Summaries: Task 2: 3 Attributing Sentences. L/Label: F=Full Support, P=Partial Support, N=No Support, U=Unclear.}
  \label{tab:crisisfacts-machine-task2}
\end{table}

\begin{adjustwidth}{-\extralength}{0cm}

\reftitle{References}


\bibliography{references_human_eval_attrib}

%


\PublishersNote{}
\end{adjustwidth}

\end{document}